\documentclass{article}
\usepackage{spconf,amsmath,graphicx}

\usepackage{amsmath}
\usepackage{amssymb}
\usepackage[numbers]{natbib}
\usepackage{xcolor}
\usepackage{setspace}
\usepackage{subcaption}
\usepackage{booktabs}

\usepackage{multirow}
\usepackage{graphicx}
\usepackage{float}

\definecolor{White}{rgb}{1.,0.,1.}
\definecolor{first}{rgb}{.8,.0,.0}
\definecolor{second}{rgb}{.0,.6,.0}
\definecolor{third}{rgb}{.0,.0,.8}

\definecolor{ceiling}{RGB}{214,  38, 40}
\definecolor{floor}{RGB}{43, 160, 4}
\definecolor{wall}{RGB}{158, 216, 229}
\definecolor{window}{RGB}{114, 158, 206}
\definecolor{chair}{RGB}{204, 204, 91}
\definecolor{bed}{RGB}{255, 186, 119}
\definecolor{sofa}{RGB}{147, 102, 188}
\definecolor{table}{RGB}{30, 119, 181}
\definecolor{tvs}{RGB}{160, 188, 33}
\definecolor{furniture}{RGB}{255, 127, 12}
\definecolor{objects}{RGB}{196, 175, 214}

\definecolor{car}{rgb}{0.39215686, 0.58823529, 0.96078431}
\definecolor{bicycle}{rgb}{0.39215686, 0.90196078, 0.96078431}
\definecolor{motorcycle}{rgb}{0.11764706, 0.23529412, 0.58823529}
\definecolor{truck}{rgb}{0.31372549, 0.11764706, 0.70588235}
\definecolor{othervehicle}{rgb}{0.39215686, 0.31372549, 0.98039216}
\definecolor{person}{rgb}{1.        , 0.11764706, 0.11764706}
\definecolor{bicyclist}{rgb}{1.        , 0.15686275, 0.78431373}
\definecolor{motorcyclist}{rgb}{0.58823529, 0.11764706, 0.35294118}
\definecolor{road}{rgb}{1.        , 0.        , 1.        }
\definecolor{parking}{rgb}{1.        , 0.58823529, 1.        }
\definecolor{sidewalk}{rgb}{0.29411765, 0.        , 0.29411765}
\definecolor{otherground}{rgb}{0.68627451, 0.        , 0.29411765}
\definecolor{building}{rgb}{1.        , 0.78431373, 0.        }
\definecolor{fence}{rgb}{1.        , 0.47058824, 0.19607843}
\definecolor{vegetation}{rgb}{0.        , 0.68627451, 0.        }
\definecolor{trunk}{rgb}{0.52941176, 0.23529412, 0.        }
\definecolor{terrain}{rgb}{0.58823529, 0.94117647, 0.31372549}
\definecolor{pole}{rgb}{1.        , 0.94117647, 0.58823529}
\definecolor{trafficsign}{rgb}{1.        , 0.        , 0.        }
\definecolor{otherstructure}{rgb}{0.98039215, 0.58823529, 0.}
\definecolor{otherobject}{rgb}{0.19607843, 1.        , 1.        }

\makeatletter

\newcommand{\car@semkitfreq}{3.92}
\newcommand{\bicycle@semkitfreq}{0.03}
\newcommand{\motorcycle@semkitfreq}{0.03}
\newcommand{\truck@semkitfreq}{0.16}
\newcommand{\othervehicle@semkitfreq}{0.20}
\newcommand{\person@semkitfreq}{0.07}
\newcommand{\bicyclist@semkitfreq}{0.07}
\newcommand{\motorcyclist@semkitfreq}{0.05}
\newcommand{\road@semkitfreq}{15.30}
\newcommand{\parking@semkitfreq}{1.12}
\newcommand{\sidewalk@semkitfreq}{11.13}
\newcommand{\otherground@semkitfreq}{0.56}
\newcommand{\building@semkitfreq}{14.1}
\newcommand{\fence@semkitfreq}{3.90}
\newcommand{\vegetation@semkitfreq}{39.3}
\newcommand{\trunk@semkitfreq}{0.51}
\newcommand{\terrain@semkitfreq}{9.17}
\newcommand{\pole@semkitfreq}{0.29}
\newcommand{\trafficsign@semkitfreq}{0.08}
\newcommand{\semkitfreq}[1]{{\csname #1@semkitfreq\endcsname}}

\newcommand{\car@sscbkitfreq}{2.85}
\newcommand{\bicycle@sscbkitfreq}{0.01}
\newcommand{\motorcycle@sscbkitfreq}{0.01}
\newcommand{\truck@sscbkitfreq}{0.16}
\newcommand{\othervehicle@sscbkitfreq}{5.75}
\newcommand{\person@sscbkitfreq}{0.02}
\newcommand{\road@sscbkitfreq}{14.98}
\newcommand{\parking@sscbkitfreq}{2.31}
\newcommand{\sidewalk@sscbkitfreq}{6.43}
\newcommand{\otherground@sscbkitfreq}{2.05}
\newcommand{\building@sscbkitfreq}{15.67}
\newcommand{\fence@sscbkitfreq}{0.96}
\newcommand{\vegetation@sscbkitfreq}{41.99}
\newcommand{\terrain@sscbkitfreq}{7.10}
\newcommand{\pole@sscbkitfreq}{0.22}
\newcommand{\trafficsign@sscbkitfreq}{0.06}
\newcommand{\otherstructure@sscbkitfreq}{4.33}
\newcommand{\otherobject@sscbkitfreq}{0.28}
\newcommand{\sscbkitfreq}[1]{{\csname #1@sscbkitfreq\endcsname}}

\title{Multi-Path Mamba Diffusion for 3D Point Cloud Semantic Scene Completion}
\name{Li Liang$^{\star}$ \qquad Naveed Akhtar$^{\dagger}$ \qquad Jordan Vice$^{\star}$ \qquad Anwaar Ulhaq $^{\ddag}$ \qquad Ajmal Saeed Mian$^{\star}$}
  
\address{$^{\star}$ The University of Western Australia,
        $^{\dagger}$ The University of Melbourne,
        $^{\ddag}$ Central Queensland University}
\begin{document}
%
\maketitle
\begin{abstract}
3D semantic scene completion plays a critical role in enabling various downstream tasks within autonomous systems by estimating both missing geometric structures and semantic information in scene data. In this work, we introduce a unique neural framework that integrates state space modeling with diffusion-based generative approaches to enhance 3D point cloud semantic scene completion. We propose \textit{MPath Mamba}  diffusion network, leveraging a skip-triple Mamba structure, dimensional decomposition residuals, and multi-directional dilations within multi-path data to efficiently process long-sequence data for feature extraction. The proposed method operates in the conditioned latent space of a variational autoencoder, enhanced by a state-space mechanism to improve 3D point cloud representation and reconstruction. Extensive experiments conducted on the SemanticKITTI benchmark demonstrate that our approach achieves highly competitive performance for 3D point cloud semantic scene completion. 
\end{abstract}
\section{Introduction}
\label{sec:intro}
3D scene understanding is a cornerstone of autonomous driving, combining geometric and semantic analysis to facilitate reliable perception and informed decision-making. LiDAR sensors play an essential role in this domain, providing high-precision, long-range measurements that surpass the capabilities of cameras in various scenarios. However, achieving accurate semantic scene completion from 3D point clouds in large-scale outdoor environments continues to pose a significant challenge. This difficulty arises from the inherent sparsity of LiDAR data and the complexity of real-world scenes containing occlusions, which hinder the development of robust and comprehensive solutions for semantic understanding.

Recent studies~\cite{song2017semantic, yan2021sparse, cheng2021s3cnet, xia2023scpnet} have significantly advanced the precision and efficiency of 3D point cloud semantic scene completion. For instance, JS3C-Net~\cite{yan2021sparse} integrates a segmentation network with a scene completion module. However, its dependence on initial segmentation introduces a potential for error propagation, which can adversely impact overall performance. S3CNet~\cite{cheng2021s3cnet} addresses this by employing sparse convolutions and a multi-view fusion strategy to enhance semantic scene completion. Similarly, SCPNet~\cite{xia2023scpnet} improves single-frame representations through dense-to-sparse knowledge distillation. However, SCPNet’s two-stage framework—comprising scene completion followed by segmentation, substantially increases the model size, a challenge further exacerbated by the absence of effective downsampling mechanisms in the technique. 

\begin{figure*}[t] 
\centering   
\includegraphics[width=0.9\textwidth]{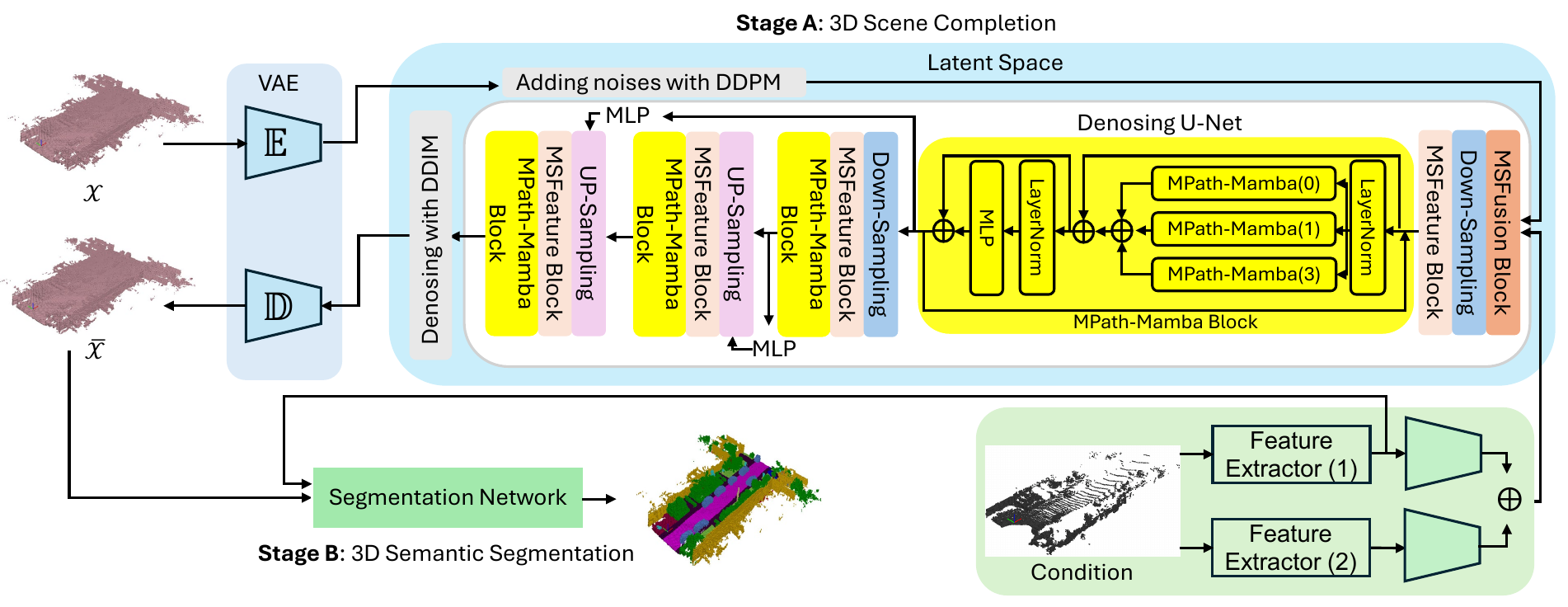}
\caption{Schematics of the proposed approach. It consists of two main components: \textbf{(A)} 3D scene completion network and \textbf{(B)} a 3D semantic segmentation network. The scene completion network operates within the latent space of a Variational Autoencoder (VAE), incorporating two conditioning sub-networks,  Multi-Scale Fusion Blocks (MSFusion), and  MPath Mamba diffusion denoising network. A modular segmentation network performs 3D semantic segmentation.}
\label{Method_1}
\end{figure*} 

This study tackles this challenge by utilizing advancements in generative diffusion modeling~\cite{rombach2022high} and state-space models~\cite{gu2023mamba}. The proposed neural network presented in Fig.~\ref{Method_1} operates within the conditioned latent space of a Variational Autoencoder (VAE). The proposed Multi-Path (MPath) Mamba Diffusion model, effectively captures rich contextual information by employing triple-directional feature dependencies with varying dilation rates within a multi-path data structure. By incorporating a combination of multi-scale fusion and feature extraction blocks with an efficient downsampling strategy, the proposed approach significantly improves computational efficiency while simultaneously improving 3D semantic scene completion performance. Through this work, we contribute:
\begin{itemize}
    \item A computationally efficient neural network design, integrating state-space and generative diffusion structures within the latent space of a VAE for 3D point cloud semantic scene completion. 
    \item The MPath Mamba diffusion model, utilizing skip-triple Mamba layers with varied dilation rates within a multi-path structure to enhance 3D scene completion.
    \item A comprehensive evaluation of our technique on the SemanticKITTI dataset demonstrates that MPath Mamba achieves highly competitive performance for 3D point cloud semantic scene completion. 
\end{itemize}
\section{Related Work}
This work contributes to 3D point cloud semantic scene completion by proposing a diffusion-based network integrated with state-space modeling. We provide a review of related work in three key domains: diffusion models, state-space models, and 3D point cloud semantic scene completion.

\noindent{\bf Diffusion Models.} Diffusion models~\cite{ho2020denoising} work on the principle of learning a sampler for the original data distribution. In the 3D domain, they have become increasingly prominent in generative tasks, such as 3D shape generation~\cite{shim2023diffusion} and 3D occupancy prediction~\cite{tang2024sparseocc}. Despite their effectiveness, these models often require high computational costs due to frequent function and gradient evaluations~\cite{patterson2021carbon}. Latent diffusion models~\cite{rombach2022high} address this challenge by performing the diffusion process within the latent space of an autoencoder, thereby significantly reducing the computational overhead. While much of the existing 3D research has centered on object-level generation, Sparseocc \cite{tang2024sparseocc} expanded this to scene synthesis by introducing a 3D sparse diffusion model that utilizes decomposed sparse kernels and a restructured transformer head.

\noindent{\bf State-Space Models.} Initially developed within the field of control theory~\cite{glasser1985control}, state-space models have proven effective in capturing long-range dependencies and have recently emerged as compelling alternatives to CNNs and Transformers in natural language processing and computer vision. In this context, S4~\cite{gu2021efficiently} utilizes diagonal parameter structures for efficient sequence modeling, which has inspired subsequent advancements such as HTTYH~\cite{gu2022train} and S4D~\cite{gu2022parameterization}, both of which strike a balance between computational efficiency and high performance. Building upon these foundations, S5~\cite{smith2022simplified} improves efficiency through parallel scanning. These developments have been further extended through the Mamba framework~\cite{gu2023mamba}, which has been successfully applied to various tasks, including point cloud processing~\cite{liu2024point}. It is noteworthy that Liu et al.~\cite{liu2024point} proposed a  point cloud backbone based on a state space model
with an octree-based ordering strategy, which is an entirely different contribution than our proposal of MPath Mamba for scene completion.  

\noindent{\bf 3D Point Cloud Semantic Scene Completion.}  3D semantic scene completion methods can be categorized into image-based~\cite{song2017semantic}, and point cloud-based~\cite{rist2021semantic}, with most relying on CNNs or transformers as their backbone. Rist \emph{et al.}~\cite{rist2021semantic} proposed Deep Implicit Functions (DIFs) using a 2D multi-resolution grid to represent the geometry and semantics of 3D scenes, while SCPNet~\cite{xia2023scpnet} leverages dense knowledge distillation and label rectification to improve single-frame representations. CNN-based methods are limited by their constrained receptive fields, whereas transformers extend receptive fields but are memory-intensive. The Mamba framework~\cite{gu2023mamba} addresses these challenges by expanding the receptive field with minimal memory overhead, making it particularly suitable for 3D point cloud semantic scene completion tasks. This is one of the key inspirations behind our unique technique.  

\section{Method}
The objective of 3D Point Cloud Semantic Scene Completion is to simultaneously infer the geometric structure and semantic information of a scene from sparse and incomplete point cloud data. The input point cloud, denoted as $P \in \mathbb{R}^{N \times 3}$, is discretized into a 3D voxel grid with dimensions $L \times W \times H$, corresponding to the spatial extent of the scene. Each voxel is assigned a semantic label $c \in {0,1,\dots, C-1}$, indicating whether it is empty or occupied by an object belonging to one of $C$ the predefined categories.

\subsection{Proposed Network}
Figure~\ref{Method_1} illustrates the proposed approach, centered on the Multi-Path Mamba (MPath Mamba) Diffusion Denoising Network. Built on a Variational Autoencoder (VAE), the framework incorporates two conditioning networks to generate compact voxel representations, reducing computational complexity while maintaining performance. A Multi-Scale Fusion (MSFusion) Block improves contextual understanding through a combination of global and local convolutional kernels and conditional features, while the MPath Mamba network integrates downsampling, multi-scale feature extraction (MSFeature), MPath Mamba block, MLP, and upsampling to achieve its objectives.

\subsection{3D Scene Completion Network}
The 3D scene completion network forms the core of our overall approach. This network is marked as Stage A in Fig.~\ref{Method_1} to delineate it from the segmentation network, which is Stage~B. The scene completion and semantic segmentation tasks are modular in our technique. 

As depicted in Fig.~\ref{Method_1}, the proposed MSFusion and MPath Mamba Diffusion Denoising Network are designed to operate within the latent space of a VAE, utilizing both the VAE and conditioning networks to enhance efficiency and performance. Inspired by \cite{milletari2016v}, the VAE is trained using a combination of cross-entropy and Lovasz-Softmax losses to reduce the blur from voxelization. For a voxelized scene $v \in L \times W \times H$, the encoder $\mathbb{E}$ maps $v$ to a latent representation $z = \mathbb{E}(v)$, and the decoder $\mathbb{D}$ reconstructs it as $\tilde{v} = \mathbb{D}(z)$. The latent space $z \in \mathbb{R}^{l \times w \times h \times c}$ is compressed by a factor of $f=4$. This 3D latent space enables efficient compression and reconstruction for the diffusion model. The conditioning networks are aligned with the encoder’s architecture for seamless integration with the VAE latent space.

The MSFusion block efficiently extracts multi-scale features by substituting large convolutional kernels with sequential  $3 \times 3 \times 3$ convolutions, drawing inspiration from SCPNet~\cite{xia2023scpnet} and VGG~\cite{simonyan2014very}. Two layers employ a $5 \times 5 \times 5$ kernel, and three layers leverage a $7 \times 7 \times 7$ kernel, reducing computational cost while maintaining capacity. For a feature map with $C$ channels, the computational cost of a $5 \times 5 \times 5$ convolution is reduced from $125C^2$ to $54C^2$, and for a $7 \times 7 \times 7$ kernel, from $343C^2$ to $81C^2$. This design efficiently captures local and global contexts, balancing performance and computational efficiency for large-scale 3D scene understanding.

The down-sampling block in the MPath Mamba network is composed of three  $3 \times 3 \times 3$ convolutional layers, each incorporating instance normalization. LeakyReLU activation is applied to the first and last layers to optimize feature extraction. A residual connection bypasses the first two layers and merges with the third, ensuring effective gradient flow and representation learning. The MLP adopts the same structure but without downsampling, supporting residue transfer between encoding and decoding. The upsampling block employs transposed convolution for resolution enhancement, followed by the downsampling block's structure for feature reconstruction. 

The MSFeature block enhances signal processing within the Mamba framework by capturing spatial relationships in 3D features before flattening them into a 1D sequence for subsequent processing. It employs Dimensional Decomposition Residual (DDR) blocks~\cite{li2019rgbd} with varying dilation rates to effectively capture scene layout information. The DDR block improves the efficiency of 3D CNN computations by decomposing convolution operations into $1 \times 1 \times k$, $1 \times k \times 1$, and $k \times 1 \times 1$ layers. This reduces computational cost from $C^{in} \times C^{out} \times k^3$ to $C^{in} \times C^{out} \times 3k$, cutting parameters by approximately one-third compared to standard $3 \times 3 \times 3$ kernels while preserving essential spatial details. 
\begin{figure}[t] 
\centering   
\includegraphics[width=0.45\textwidth]{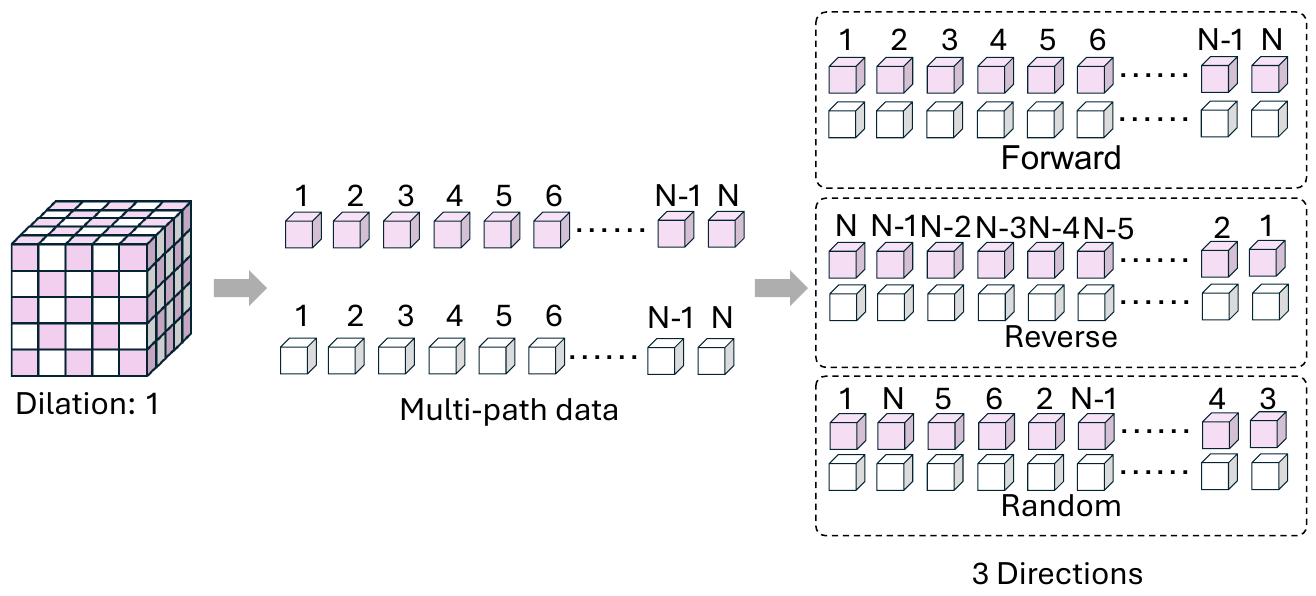}
\caption{MPath Mamba flow. Data is divided into multiple paths based on dilation rates, with each path providing three directional components as input to MPath Mamba.}
\label{Method_2}
\end{figure} 
\begin{table*}[!htbp]
\normalsize
	\newcommand{\clsname}[2]{
		\rotatebox{90}{
			\hspace{-6pt}
			\textcolor{#2}{$\blacksquare$}
			\hspace{-6pt}
			\renewcommand\arraystretch{0.6}
			\begin{tabular}{l}
				#1                                      \\
				\hspace{-4pt} ~\tiny(\semkitfreq{#2}\%) \\
			\end{tabular}
	}}
	\centering
	\caption{Quantitative results on the SemanticKITTI~\cite{SemanticKITTI} test set are from the online server. "Param." represents parameters in millions, and bold indicates the best results in each column.}
	\resizebox{\linewidth}{!}
	{
		\begin{tabular}{l|cc|c|cccccccccccccccccccc}
			\toprule		
			Method 								 & 
			IoU 								 & 
			mIoU  								 & 
                Param.  								 & 
            \clsname{car}{car} 					 & 
            \clsname{bicycle}{bicycle}           &
            \clsname{motorcycle}{motorcycle}     &
            \clsname{truck}{truck}               &
            \clsname{other-veh.}{othervehicle}   &
            \clsname{person}{person}             &
            \clsname{bicyclist}{bicyclist}       &
            \clsname{motorcyclist}{motorcyclist} &
			\clsname{road}{road}                 & 
            \clsname{parking}{parking}           & 
			\clsname{sidewalk}{sidewalk}         &  
			\clsname{other-grnd.}{otherground}   & 
			\clsname{building}{building}         & 
            \clsname{fence}{fence}               &
			\clsname{vegetation}{vegetation}     &
			\clsname{trunk}{trunk}               &
			\clsname{terrain}{terrain}           &
            \clsname{pole}{pole}                 &
            \clsname{traf.-sign}{trafficsign}
			
			\\
			\midrule   
   
            LMSCNet~\cite{roldao2020lmscnet}      & 56.7          & 17.6     &0.4M      & 30.9          & 0.0          &  0.0          & 1.5           & 0.8          &  0.0          & 0.0          &  0.0          & 64.8          & 29.0          & 34.7          & 4.6           & 38.1          & 21.3          & 41.3         & 19.9          & 32.1           & 15.0          &  0.8          \\
		UDNet~\cite{zou2021up}      & 59.4 & 19.5 &- & 33.9 & 0.8 & 0.4 & 3.8
			&  4.4 & 0.5 &  0.3 & 0.3 & 62.0 & 28.2  &  35.1 & 9.1  &  39.5 & 24.4  & 40.9
			&  23.2  & 32.3 & 18.8 &  13.1 		  \\
		Local-DIFs~\cite{rist2021semantic}    & 57.7 & 22.7 &- & 34.8 & 3.6 & 2.4 & 4.4 
			& 4.8 & 2.5 & 1.1 & 0.0 &  67.9 & 40.1  & 42.9 & 11.4  &  40.4 &  29.0  & 42.2
			& 26.5  & 39.1 &  21.3 &  17.5         \\
		SSA-SC~\cite{yang2021semantic}      & 58.8 & 23.5  &41.0M & 36.5 & 13.9 & 4.6 &  5.7      
			&  7.4 &  4.4 & 2.6 &  0.7 & 72.2 &  37.4  & 43.7 & 10.9  &  43.6 &  30.7  & 43.5
			& 25.6  & 41.8 &  14.5 &  6.9        \\
		JS3C-Net~\cite{yan2021sparse}   & 56.6 & 23.8 &3.1M & 33.3 & 14.4 & 8.8 & 7.2 & 12.7 & 8.0 & 5.1 & 0.4 &  64.7 & 34.9 &  39.9 & 14.1 &  39.4 & 30.4 & 43.1 &  19.6 & 40.5 & 18.9 &  15.9 \\
            SSC-RS~\cite{mei2023ssc} & 59.7 & 24.2 &23.0M &36.4 &  10.1 & 5.1 &  5.3 & 11.2 & 4.7 &  2.4 &  0.9 & 73.1 & 38.6 & 44.4 &17.4 &44.6 &  30.8 & 44.1 & 26.0 & 41.9 &  15.0 &  7.2 \\
		S3CNet~\cite{cheng2021s3cnet}     & 45.6 & 29.5 &- & 31.2 & 41.5 & 45.0 & 6.7
			& 16.1 & 45.9 & 35.8 & 16.0 & 42.0 & 17.0  &  22.5 &  7.9  & 52.2 & 31.3  & 39.5 
			&  34.0  & 21.2 &  31.0 & 24.3         \\
   		\hline
            Ours & 55.6 & 29.0 & 29.4M & \textbf{40.1} &  22.5 & 19.8 & \textbf{15.5} & \textbf{20.4} & 14.2 & 12.9 &  0.5 & 70.9 & \textbf{45.7} & \textbf{45.5} & \textbf{25.2} & 36.4 & \textbf{35.5} & 43.1 & 33.7 & \textbf{45.2} & 19.6 & 4.4  \\
			\bottomrule
		\end{tabular}
	}
	\setlength{\abovecaptionskip}{0cm}
	\setlength{\belowcaptionskip}{0cm}
	\label{tab:sem_kitti_test}
\end{table*}

The proposed MPath Mamba Block is a key component of the MPath Mamba Denoiser, utilizing a Triple Mamba configuration inspired by \cite{yang2024vivim} to capture both direct and indirect feature connections, enhancing contextual information. It incorporates MPath Skip Triple Mamba (MSTM) layers with varying dilation rates across forward, reverse, and random directions within a multi-path structure. As shown in Fig.~\ref{Method_2}, an example with a dilation rate of 1 demonstrates how each MSTM layer calculates feature dependencies; \textcolor{black}{the same procedure is applicable when dilation rates of 0 and 3 are employed.} 
effectively capturing comprehensive feature relationships. The output of the $i^{\text{th}}$ MSTM layer is defined as: 
\begin{align}
&\Omega_{i}  = \Omega_{i}^{f}(z_{d}) + \Omega_{i}^{r}(z_{d}) + \Omega_{i}^{s}(z_{d}), \\
&\Omega_{MPath} = \sum_{i} \Omega_{i}, i \in \left \{0, 1, 3 \right \},
\end{align}
where $z_{d}$ represents the input multi-path features, while $f$, $r$, and $s$ denote the forward, reverse, and random directions, respectively. The block's final output, $\Omega_{final}$, is computed as  $\Omega_{final} = \text{MLP}(\text{LN}(\Theta_{in})) + \Theta_{in}$, where $\Theta_{in} = \Omega_{MPath}(\text{LN}(\phi_{\text{initial}})) + \phi_{\text{initial}}$. Here, $\phi_{\text{initial}}$ represents the input features of the MPath Mamba block, and MLP and LN refer to Multi-Layer Perceptrons and Layer Normalization. By utilizing varying dilation rates across the three directions within multi-path data, the block captures diverse feature dependencies, enhancing connectivity and improving performance in 3D semantic scene completion.
\subsection{3D Semantic Segmentation Network}
The 3D Semantic Segmentation Network is designed to process and segment the outputs from the 3D completion network. Its architecture incorporates multi-path blocks inspired by SCPNet~\cite{xia2023scpnet} and DDR blocks~\cite{li2019rgbd}, enabling efficient capture of both local and global contextual features. The network is designed to process and segment the outputs from the 3D completion network. Its architecture incorporates multi-path blocks inspired by SCPNet~\cite{xia2023scpnet} and DDR blocks~\cite{li2019rgbd}, enabling efficient capture of both local and global contextual features. \textcolor{black}{In addition, feature extractor-(1) employs multi-path blocks~\cite{xia2023scpnet} to extract robust 3D spatial and contextual information, while feature extractor-(2) utilizes a 3D feature extraction network that incorporates multi-path blocks \cite{xia2023scpnet} and DDR blocks \cite{li2019rgbd}, along with a 2D image feature extraction network \cite{yang2021semantic}. This combined approach integrates 3D spatial features and 2D spatial features to provide more robust contextual representations.}
\subsection{Training Objective}
The training objective, designed for the complexity of the task, uses a composite loss function. The MPath Mamba denoising network follows the diffusion denoising framework, minimizing the expected squared error between the predicted output and target data, expressed as:
\begin{align}
    L_{DM} = \mathbb{E}_{x,\epsilon \sim N\left ( 0, 1 \right ) ,t}\left [ \left \| \epsilon - \epsilon _{\theta}\left ( x_{t}, t  \right )   \right \| _{2}^{2}   \right ],
\end{align}
where $\epsilon _{\theta}\left ( x_{t}, t  \right )$ represents an equally weighted sequence of denoising autoencoders, with $t\in \{1,2,..,T\}$. The model is trained to predict the denoised version of the input $x_{t}$, which corresponds to the noisy version of the original input $x$ at time step $t$. For 3D semantic segmentation, the objective function is defined as:
\begin{align}
\mathcal{L} &= \mathcal{L}_{CE} + \beta \mathcal{L}_{Lovasz},
\end{align}
where the balancing coefficient $\beta$,  which determines the relative contributions of different loss components, is set to 1 in the proposed method. The cross-entropy loss $\mathcal{L}_{CE}$ and Lovasz-Softmax loss $\mathcal{L}_{Lovasz}$ are defined as:
\begin{align}
\mathcal{L}_{CE} = - \sum_{i} y_{i} \log(\hat{y_{i}}),~~
\mathcal{L}_{Lovasz} = \frac{1}{|C|} \sum_{c=1}^{C} J \left ( e(c) \right ),
\end{align}
where $\hat{y}_i$ and $y_{i}$ represent the predicted output and ground truth for the $i^{\text{th}}$ element, $J$  is the Lovasz extension of IoU, and $e(c)$ is the error vector for class $C$ in the set of classes. The VAE objective combines $\mathcal{L}_{CE}$, $\mathcal{L}_{Lovasz}$, and the Kullback-Leibler (KL) divergence to ensure effective reconstruction and regularization.
\begin{figure*}[t]
    \centering
    \begin{subfigure}{\textwidth} 
        \centering   
        \includegraphics[width=0.9\textwidth]{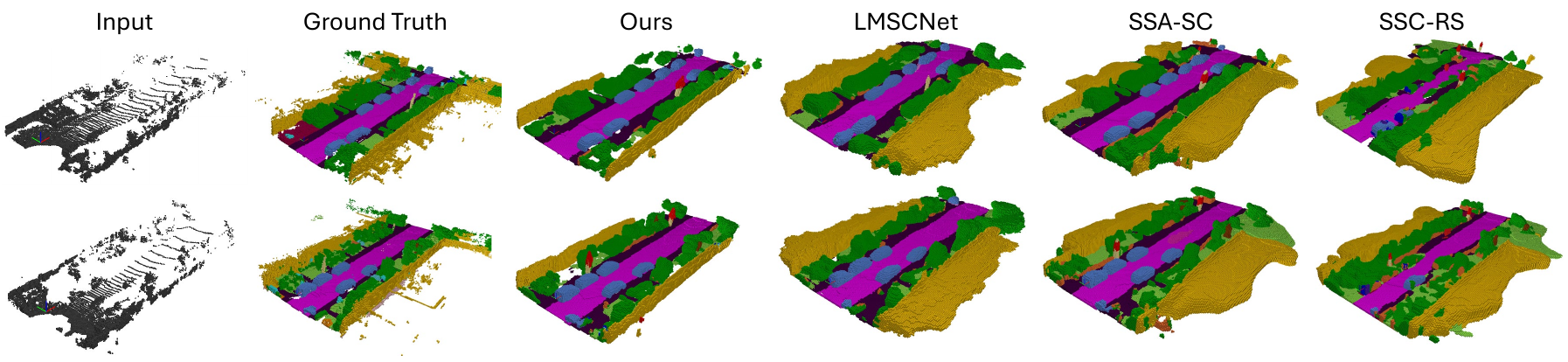}
    \end{subfigure}
    \vspace{0.5mm} 
    \begin{subfigure}{\textwidth} 
        \centering   
        \includegraphics[width=\textwidth]{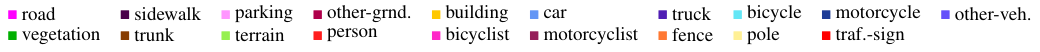}
    \end{subfigure}
    \caption{Qualitative results on the SemanticKITTI validation set. Columns from the left represent input data, ground truth, and outputs of our method, LMSCNet, SSA-SC, and SSC-RS.}
    \label{Comparison}
\end{figure*}
\section{Experiments}
The proposed network is evaluated on the SemanticKITTI dataset~\cite{behley2019semantickitti}, a widely recognized benchmark for 3D semantic scene completion. \textcolor{black}{This is the only outdoor dataset used to evaluate existing 3D point cloud semantic scene completion methods.} Ablation studies are also performed to assess the contributions of each component of our framework.\\
\noindent \textbf{Datasets.} The SemanticKITTI benchmark~\cite{behley2019semantickitti} provides densely annotated urban driving scenes. The dataset includes voxelized scenes within a $51.2m \times 51.2m \times 6.4m$ volume, represented as $256 \times 256 \times 32$ voxel grids with a resolution of 0.2 m. It comprises 10 training sequences, 1 validation sequence, and 11 testing sequences, covering 20 classes.\\
\noindent \textbf{Evaluation Metrics.} The proposed framework is evaluated using standard metrics for semantic scene completion~\cite{rist2021semantic}, focusing on both geometric reconstruction and semantic segmentation accuracy. Geometric performance is assessed using Intersection over Union (IoU), defined as $IoU_{i} = \frac{TP_{i}}{TP_{i} + FN_{i} + FP_{i}}$
 , where $TP_i$, $FP_i$, and $FN_i$ denote the true positives, false positives, and false negatives for the $i^{\text{th}}$ class, respectively. Semantic segmentation accuracy is evaluated across all classes using Mean Intersection over Union (mIoU), calculated as $mIoU = \frac{1}{C} \sum_{c=1}^{C} IoU_{c}$, where $C$ represents the total number of classes.
\subsection{Results}
The proposed method is compared against several state-of-the-art approaches in semantic scene completion, \textcolor{black}{including LMSCNet~\cite{roldao2020lmscnet}, UDNet~\cite{zou2021up}, Local-DIFs~\cite{rist2021semantic}, SSA-SC~\cite{yang2021semantic}, JS3C-Net~\cite{yan2021sparse}, SSC-RS~\cite{mei2023ssc}, and S3CNet~\cite{cheng2021s3cnet}.} 

Table~\ref{tab:sem_kitti_test} shows the Semantic-KITTI test results for 3D point cloud semantic scene completion. The proposed method achieves competitive performance, surpassing most state-of-the-art methods for mIoU, except S3CNet~\cite{cheng2021s3cnet}. Our method performs comparably to S3CNet in segmentation while outperforming it considerably in completion. We note that another method, 
 SCPNet~\cite{xia2023scpnet}, also reports highly competitive results. However, that method also entails a very high computational cost that far exceeds our NVIDIA 4090 GPU resources.  
Our method ensures robust performance across diverse object scales, excelling in both large-scale (e.g., parking) and small-scale objects (e.g., cars, trucks) while keeping the compute requirements reasonable. 

We also utilize the pre-trained models and official implementations of LMSCNet~\cite{roldao2020lmscnet}, SSA-SC~\cite{yang2021semantic}, and SSC-RS~\cite{mei2023ssc} for qualitative comparisons on the SemanticKITTI validation set. As illustrated in Fig.~\ref{Comparison}, the proposed method demonstrates excellent performance in segmenting planar categories (e.g., roads, terrain) and consistently provides accurate scene completions, effectively avoiding the distorted reconstructions observed in the compared methods.
\subsection{Ablation Study}
Systematic experiments are conducted to evaluate the impact of various components within the proposed framework, as detailed in Table~\ref{tab:sem_kitti_val_ablation}. Performance variations were analyzed by removing specific blocks from the framework. The ablation study underscores the significance of MSFusion and MPath Mamba in improving the performance of 3D semantic scene completion. The results demonstrate that the integration of components is crucial for capturing complex spatial and semantic relationships. The configuration without MPath Mamba denotes the use of a Triple Mamba block.
\section{Conclusion}
In this paper, we propose a multi-path Mamba denoising diffusion network for efficient and robust 3D point cloud semantic scene completion. The proposed MPath Mamba network combines variational autoencoders with dual conditioning to generate compact latent representations, ensuring efficiency and performance. For this purpose, it employs skip triple dilations to capture direct and indirect feature relationships within multi-path data, resulting in the enhancement of spatial and semantic representations. Evaluations on the SemanticKITTI dataset validate the effectiveness and robustness of the approach.
\begin{table}[t]{
	\newcommand{\clsname}[2]{
		\rotatebox{90}{
			\hspace{-6pt}
			\textcolor{#2}{$\blacksquare$}
			\hspace{-6pt}
			\renewcommand\arraystretch{0.6}
			\begin{tabular}{l}
				#1                                      \\
				\hspace{-4pt} ~\tiny(\semkitfreq{#2}\%) \\
			\end{tabular}
	}}
        \vspace{-1cm}
	\centering
	\caption{Ablation study on the SemanticKITTI validation set for 3D semantic Scene Completion (IoU and mIoU). w/o $<$MPath Mamba$>$ indicates the use of 
    $<$Triple Mamba$>$.}
    \resizebox{\linewidth}{!}
	{    
		\begin{tabular}{l|cc}
			\toprule		
			Method								 & 
			\multicolumn{2}{c}{3D Semantic Scene Completion}                  
             \\
                                                     &
                IoU                                  &
                mIoU
                \\
			\midrule
             w/o MSFusion   & 46.5 & 26.2 \\
             w/o MSFeature  & 48.7 & 28.4 \\
             w/o MPath Mamba & 49.4 & 27.8 \\
             Full Model  & 51.2 & 29.1 \\
			\bottomrule
		\end{tabular}
	}
        \vspace{-0.5cm} 
	\setlength{\abovecaptionskip}{0cm}
	\setlength{\belowcaptionskip}{0cm}
	\label{tab:sem_kitti_val_ablation}}
\end{table}

\section{Acknowledgments}
Dr. Naveed Akhtar holds an ARC Discovery Early Career Researcher Award (DE230101058), and Professor Ajmal Saeed Mian holds an ARC Future Fellowship (FT210100268), both funded by the Australian Government.
\bibliographystyle{IEEEbib}
\begin{spacing}{0.2}
\bibliography{Li}
\end{spacing}
\end{document}